\documentclass{article}
\usepackage{ijcai17}
\usepackage{graphicx}
\usepackage{times}
\usepackage{mathtools}
\usepackage{amsfonts}
\usepackage{mathrsfs}
\usepackage{algorithm}
\usepackage{algorithmic}
\usepackage{multirow}
\usepackage{caption}
\usepackage{booktabs}
\usepackage{float,dblfloatfix}
\usepackage{url}
\title{Open-Category Classification by Adversarial Sample Generation\thanks{This research was supported by the NSFC (61375061, 61333014), Jiangsu SF (BK20160066), and Foundation for the Author of National Excellent Doctoral Dissertation of China (201451).}}
\author{
Yang Yu$^{1}$, 
Wei-Yang Qu$^1$, 
Nan Li$^2$, 
Zimin Guo$^3$\thanks{Part of the work was done when Z. Guo was visiting Nanjing University as an undergraduate student.}\\
$^1$National Key Laboratory for Novel Software Technology, Nanjing University, Nanjing, China\\
$^1$Collaborative Innovation Center of Novel Software Technology and Industrialization, Nanjing, China\\
$^2$Alibaba Group, Hangzhou, China\\
$^3$College of Engineering, UC Berkeley\\
yuy@nju.edu.cn
}

\begin{document}

\maketitle

\begin{abstract}
	In real-world classification tasks, it is difficult to collect training samples from all possible categories of the environment. Therefore, when an instance of an unseen class appears in the prediction stage, a robust classifier should be able to tell that it is from an unseen class, instead of classifying it to be any known category. In this paper, adopting the idea of adversarial learning, we propose the ASG framework for open-category classification. ASG generates positive and negative samples of seen categories in the unsupervised manner via an adversarial learning strategy. With the generated samples, ASG then learns to tell seen from unseen in the supervised manner. Experiments performed on several datasets show the effectiveness of ASG.
\end{abstract}

\section{Introduction}
As machine learning techniques are adopted in increasing applications, it is appealing that they can be applied in environments that are open and non-stationary, where unseen situations can emerge unexpectedly. For classification, a typical learning task, classical methods implicitly assume that the data is i.i.d. even for the future test ones. This assumption no longer holds in open environments, which drastically weaken the robustness of classical classification methods.% face great challenges of robustness in these environments. 

In this work, we consider the open-category classification (OCC) problem, where there are novel classes that none of their instances were observed during the training phase, but in the test phase their instances could be encountered. Classical approaches can only predict instances from unseen classes as one of the seen classes. An open-environment aware classifier, on the contrary, should be able to tell at first if an instance belongs to a seen class. 

Different directions have been explored related to the OCC problem. %, including class incremental-learning, learning with a rejection option, outlier detection, zero-shot learning, etc. For examples, 
In class incremental-learning~\cite{fink2006online,Muhlbaier09learn,Kuzborskij13}, new classes are assumed to appear incremental. However, these studies mainly focused on how to enable the system to incorporate later coming training instances from new classes, but did not address the problem of recognizing unseen classes. In learning with a rejection option~\cite{chow1970optimum}, the classifier rejects to recognize an instance if its confidence is low, %. One may utilize the rejection option to determine unseen class instances, 
which, however, does not take open-environment into consideration. Note that an instance close to the seen class boundary can have a low confidence but belongs to a seen class, while an unseen class instance can have a high confidence if it is far from the seen class boundary. Outlier detection techniques~\cite{hodge2004survey} could be employed if we treat unseen class instances as outliers. However, %outlier detection addresses a different problem than the OCC problem. S
samples from seen class could also contains outliers, meanwhile, a cluster of unseen class instances may not be treated as outliers. Zero-shot learning~\cite{Palatucci09zeroshot} is close to the OCC problem, but they commonly assume that a high-level attribute set is available for all classes including the unseen one, which provides useful information for recognizing unseen class instances. In our problem, however, we consider learning in an unfamiliar open environment and thus do not assume the availability of such high-level attributes.

Only a few previous studies addressed the OCC problem. For examples, in \cite{scheirer2013toward} the open-set cost is considered, but it is hard to measure that cost without seeing open-set data; in \cite{Da2014lacu}, a set of unlabeled data of all classes is employed, but sufficient unlabeled data from all potential classes may not always available. We are thus more interested in solving the problem without extra information. 

% The key challenges of the task lies in that no instances of novel classes
% were observed during the training stage, which means that it is impossible to
% directly use the traditional supervised learning method to address this
% problem. An effective approach is to determine the boundaries for each
% category separately, with the boundaries, when an instance appears within the
% boundary of a class, the instance is considered to belong to this class,
% otherwise it does not belong to the class. According to this idea, when an
% instance does not belong to any of the seen categories, it is considered an
% instance of novel class. So the question is translated into how to determine
% the appropriate boundary for each category. Considering that, as a
% classification task, the usual practice is to establish a model first, and
% then use the model to determine whether the instance belongs to a category.
% Which means that once the model is set up, it corresponds to a classification
% boundary. So the problem turns into how to build a suitable model, note that
% if only the training data is used to build the model, then the model divides
% the feature space into fixed classes. For example, using only three categories
% of \textit{bird},\textit{cat} and \textit{fish} for training, the feature
% space will be divided into three parts, which will make the boundaries of each
% class too large. If the model sees a novel class of instance, such as a dog,
% the instance is likely to be predicted as \textit{cat}, which means the model
% can not be applied in practice.

In this paper, we propose the adversarial sample generation (ASG) framework for the OCC problem. Inspired by the adversarial learning~\cite{Goodfellow2014gan}, ASG generates negative instances of seen classes by finding data points that are close to the training instances, given that they can be separated from the seen data by a discriminator. When the training data is small, ASG also generates positive instances that cannot be discriminated from the seen class instances, in order to enlarge the dataset. Using the supervision of the generated negative and positive samples of seen classes, it is then straightforward to train an open-category classifier to tell seen from unseen. We conduct experiments on several domains with open categories but no extra training information. The results show that the ASG achieves significant better performance than the compared methods. %The contributions of this paper include\\[3pt]
%$\bullet$ A data generation method using the adversarial learning principle is proposed;\\
%$\bullet$ An open-category classification approach is proposed using the adversarial data generation method;\\
%$\bullet$ The proposed approach, learning in open-category domains with no extra information achieves outstanding performance.

The rest of this paper is organized in four sections that presents the background, the proposed method, the experiment results, and the conclusion, respectively.

% as well as unseen classes by  As discussed above, in order to determine the boundary of each seen
% category more precisely, only the training instances are not enough, we need
% additional instances as the model input, so the problem is further translated
% into how to get the additional instances. To solve this problem, we propose
% the ASG framework. In this framework, samples are first generated separately
% for each seen category, at this point, a good option is to use the GAN model
% \cite{Goodfellow2014gan}. With the generated samples, the second step is to select
% them to obtain the samples that are useful for determining the classification
% boundaries details will be in section 3. It is worth mentioning that in the
% process of selecting
% samples, it is difficult to use gradient descent to optimize the objective
% function, so we use the RACOS\cite{Yu2016racos}. method for the optimization. Once
% the bounds of each seen class are identified, when a test sample is drawn, if
% the instance is within the bounds of a seen category, it is determined that it
% belongs to this class, otherwise it does not belong to. If an instance does not
% belong to any seen category, it is considered an instance of novel category.
% The experiments on the data sets show that our method has a better performance
% than the contrast methods.

\section{Background}

\subsection{Related Problems}

This paper studies the open-category classification problem, where no information about the unseen classes, neither instances nor attributes, is available. Some studies are related to this problem.

The \textit{incremental learning} requires a proper adaptation of traditional machine learning approaches to deal with the dynamic and open environment, in addition, class-incremental learning(C-IL)~\cite{zhou2002hybrid} is an important branch of it, which mainly concerned with the addition of new classes. In \cite{fink2006online,Kuzborskij13}, each new class has a binary classifier which distinguishes between existing categories and new categories, and the new 
category in the classifier shares the hypothesis with the existing category to train. However, the method still need a few instances of new categories, when no instances of new categories are available in the training phase, these methods can not be applied. To address this limitation, a new method is introduced in \cite{Da2014lacu}, which processes new categories by exploring unlabeled instances, this requires reliable unlabeled data, but the data availability and quality is difficult to guarantee in practice.

The \textit{open set recognition} problem is mainly concerned by the community of pattern recognition, and has been applied to face recognition and speech recognition, etc. In \cite{phillips2011evaluation}, the
main concern is the operation of the threshold, only instances where the confidence is above the threshold are classified as seen classes. In \cite{scheirer2013toward,Bendale15openworld}, both the new decision boundary and the risk over open space are considered to limit the regions for seen categories.

The \textit{outlier detection} problem \cite{hodge2004survey} requires the identification of anomaly instances from a given data set. The outlier detection methods can be applied in open-category problems as long as the abnormal instances is predicted as novel classes. However, the difference is that outlier detection is only concerned with the discovery of abnormal instances, which is limited to specific applications. Besides, it does not take into account the classification error of existing categories

The \textit{class discovery} problem tries to identify the instances of rare categories which are not known in advance, but are known to exist in the training data~\cite{pelleg2004active,hospedales2013finding}. The open-category problem is different from class discovery problem, because the unseen class is not necessarily a rare class. On the other hand, it is possible to find examples of a rare class in the training data, while the instance of novel
classes only appear in the test data.

The \textit{zero-shot} problem tries to identify the unseen classes with the assumption
that some high-level attributes of all classes are known as a prior.  For example, in \cite{xian16latentembeding}, a latent embedding model was presented  which learns
a compatibility function in the high-level attributes space considered between image and class embeddings.
In this work, we do not assume the availability of such information.

\subsection{Adversarial Learning}
The adversarial learning employs a generative model and a discriminative model, where the generative model learns to generate instances that can fool the discriminative model as a non-generated instance, which is also called Turing Learning in \cite{Li16turinglearning}. Besides, the Generative adversarial nets (GAN) combines the two models as a whole neural network for end-to-end training, resulting in a model that is consistent with the original data distribution. Further improvements include studies on the stability of GAN. For example, in~\cite{Nowozin2016fgan}, the
F-divergence is introduced from the perspective of distance measurement, and GAN
has been shown to be a special case of F-divergence when it comes to a particular
metric. 
%In \cite{chen2016infogan}, the info-GAN model has been introduced and the model
%successfully acquires a disentangled representation of the data and has a
%stable training effect. In application, \cite{Dosovitskiy2016PSgan} introduces the
%perceptual similarity, it changes the past in accordance with the image
%pixel-level differences to measure the loss of the situation, which makes the model more robust. 

In our work, we will employ the adversarial learning principle that a generative model fights to generate instances judged by a discriminator. Different with the studies on GAN, our framework can apply to various learning models besides neural networks. Moreover, to solve the open-category classification problem, we do not only need to generate seen class data, but more importantly need to generate unseen class data.

\subsection{Derivative-free Optimization}

Previously learning approaches commonly employed gradient-based optimization methods. However, the optimization problems may not always simple enough to fit the gradient-based methods. Often, a complex optimization has to be relaxed to a convex problem, sacrificing the faithfulness to the original problem.

Ancient derivative-free optimization methods include representatives such as genetic algorithms \cite{golberg1989genetic}, which are mostly heuristic methods. Recently, the derivative-free optimization methods have made significant progress in both theoretical foundation and practical usage,  including Bayesian optimization methods \cite{brochu2010tutorial}, optimistic optimization methods \cite{munos2014bandits}, and  model-based optimization \cite{Yu2016racos}. A derivative-free optimization method considers an optimization formalized as $\arg\max_{x \in X} f(x)$, where $X$ is domain. The method, instead of calculating gradients of $f$, samples solutions $x$ and learns from their feedbacks $f(x)$ for finding better solutions. Therefore, derivative-free optimization methods can be more suitable for problem with bad mathematical properties, including non-convexity, non-differentiability, and having many local optima. We thus employ the state-of-the-art derivative-free methods in our approach.

\section{Proposed Method}

The open-category classification (OCC) problem can be described as follows. Given a training dataset $D = \{(x_i,y_i)\}_{i=1}^L$, where $x_i \in R^d $ is a training instance and $ y_i \in Y = \{1,2, ..., K\}$ is the corresponding category label. In the test phase, we need to predict the categories of an open dataset $D_o = \{(x_i,y_i)\}_{i=1}^\infty $, where $ y_i \in Y_o = \{1,2,...,K,K+1,...,M \} $ with  $ M > K$. Since there are classes which are not observed in the training phase, the goal of OCC is to learn a model $f(x) : X \to Y' = \{1,2,...,K,novel\}$, where the option $novel$ indicates that the category was unseen in the
training phase, so as to minimize the expected risk as follows
\begin{equation}\label{eq1}
	f^* = \mathop{\arg\min}\nolimits_{f\in\mathcal{H}}\mathbb{E}_{(x,y)\sim D_o}err(y,f(x))
\end{equation}
where $\mathcal{H}$ is the hypnosis space and $err$ is defined as follows
\begin{equation}\label{eq2}
	err(y,f(x)) =
	\begin{cases}
		I(f(x)\ne y),      & y \in Y    \\
		I(f(x) \ne novel), & y\not\in Y
	\end{cases}
\end{equation}
The $I$(expression) is an indicator function, it equals 1 when the
expression holds and 0 otherwise.

The OCC problem is difficult to solve because no information about the $novel$ class is available. Our overall idea is that, % , the feature space is divided into specific $K$ parts, when an instance of new category comes, the model can only predict it in these $K$ parts, which is inappropriate. However, if we can transform the OCC problem into traditional machine learning classification problem, the previous machine learning methods can be applied. Note that there are $K+1$ classes in the OCC problem, denoted as $Y' = \{1,2,...,K,novel\}$ . But the instance of category $novel$ is missing in the training set, which is the difficulty of the problem. 
if we can generate the instances of the class $novel$ and put them into the training set (denoted the augmented training set as $\tilde{D}$), then the problem will be easily
solved by standard supervised learning
\begin{equation}\label{eq3}
	f^* = \mathop{\arg\min}\nolimits_{f\in\mathcal{H}}\mathbb{E}_{(x,y)\sim \tilde{D}}I(y \ne f(x))
\end{equation}
The label
of instance $x$ can be predicted as $ \mathop{\arg\max}\nolimits_{k=1,...,K,novel} f_k(x)$. The problem is then how to generate data of the $novel$ class.

The idea of adversarial learning~\cite{Goodfellow2014gan} is employed for data generation. In the adversarial learning, a generative model is trained to generate samples that are thought to be appropriate according to a discriminator. For example, to generate data consisting with the training data, the objective is that the discriminator cannot distinguish the generated data from the training data. Using this idea, we directly generate the data of unseen class for the OCC problem.
\vspace{5pt}
 
\noindent{\bf Generate Unseen Class Instances}\\
To generate an instance of the unseen classes, ASG searches in the instance space, such that the instance should not be recognized as the seen class by the discriminator, which is a classifier trained to separate generated samples and seen class instances. However, there are too many such instances. For the purpose of distinguishing seen from unseen, we only need the samples that are around the boundary between seen and unseen classes. Therefore, ASG tries to find an instance that is close to the seen class instances, but is recognized as unseen class by the discriminator.

ASG considers each class separately, and for each class, it generates samples one by one. Let $P_D(x; D, D^-)$ denote the probability of $x$ to be positive by the discriminator, trained with positive data $D$ and negative data $D^-$. For class $k$, denote the current generated samples as $D_k^-$, which is empty initially. The objective that a generated sample does not belong to the seen class is 
%ASG tries to produce one sample that is not 
%samples that are not the same, denoted as $D_k^-$.
%Given the data in training set of class $k$ as $D_k$, the optimization objective is 
\begin{equation}\label{eq5}
	\mathop{\arg\min}\nolimits_{x} P_D(x; D_k, D_k^-\cup\{x\})
\end{equation}
Intuitively, we evaluate a generated sample by adding it to the negative data set and train the model to see if the generated sample is not classified as positive (seen class).

Eq.\eqref{eq5} alone cannot generate all the boundary samples of the seen class, but only data samples that do not belong to the seen class. To generate boundary samples, we further require that the generated samples are close to the seen class data. Therefore, it is natural to consider that the distance between the generated sample and the original data set $D_k$ should be small in a measure, which is enforced by the penalty term as
\begin{equation}\label {eq6}
	P_1(x,D_k)=\max\{0,\mathop{\mathop{\arg\min}}\nolimits_{x' \in D_k}dist(x,x')-C_1\}
\end{equation}
where the radius parameter $C_1$ is a positive constant,
and $dist(x,x')$ is a distance measure that can be Euclidean distance or other distance measures.

\begin{algorithm}[t]
	\caption{Generation of negative instances of seen classes}
	\label{alg:ASG}
	%\hspace*{0.02in} {\textbf {Input}:}
	\begin{algorithmic}[1]
		\REQUIRE ~~\\
		$D$: Training instances$\{(x_i,y_i)\}_{i=1}^L$ \\
		$T$: Number of generated instances per class\\
		$\mathcal L$: Learning algorithm for the discriminant model\\
		$Opt$: A derivative-free optimization method
		\ENSURE ~~\\
		$D^-$: Negative samples of all class {$1,2,..K$}
		\FOR {each $k \in \{1,...,K\}$}
		\STATE $D_k^- = \emptyset$
		\FOR {$t = 1,2,...,T $}
		\STATE  $x^-$ = solve Eq.\eqref{eq9} by $Opt$ with discriminator $\mathcal L$ 
		\STATE  Update $ D_k^- = D_k^- \cup \{x^-\} $
		\ENDFOR
		\ENDFOR
		\RETURN {$D^-=\{D_1^-, D_2^-, \ldots, D_K^-\}$}
	\end{algorithmic}
\end{algorithm}

%From the  Eq.\eqref{eq6} can be seen, the smaller the better punishment items.
%So the optimization objective function becomes
%\begin{equation}\label{eq7}
%	\mathop{\arg\min}_{x}1 - \ell(x,D_k)+\lambda_1P_1(x,D_k)
%\end{equation}
%where $ \lambda_1$ is the coefficient of $P_1$.

Furthermore, since ASG generates samples one by one, Eq.\eqref{eq5} and Eq.\eqref{eq6} cannot prevent it from generating many identical samples. However, we want the generated samples to be scattered around the boundary. Therefore, we force the generated samples to be different, i.e., a newly generated sample should be far away from the previously generated samples, which is expressed as
\begin{equation}\label{eq8}
	P_2(x,D_k^-)=\max\{0,C_2-\mathop{\mathop{\arg\min}}\nolimits_{x' \in D_k^-}dist(x,x')\}
\end{equation}
Here the radius parameter $C_2$ is a positive constant.

Combining the loss and the penalty terms, the overall objective function is
\begin{equation}\label{eq9}
	\mathop{\arg\min}_{x}  P_D(x; D_k, D_k^-\cup\{x\}) +\lambda_1 P_1(x,D_k)+\lambda_2 P_2(x,D_k^-)
\end{equation}
where $\lambda_1$ and $\lambda_2$ are hyper-parameters of the penalty terms. Note that our penalty terms aims only at are pushing samples out of the radius distance, the hyper-parameters will not have a great impact on the learning result.

Note that Eq.\eqref{eq9} is non-convex, particularly when models other than neural networks are considered. It is hard to rely on the gradient-based method. Thus, we employ the derivative-free method to solve the optimization. Since most derivative-free methods are generally applicable, we denote such an algorithm as $Opt$ in Algorithm 1. The concrete algorithm will be disclosed in the experiment section.

Algorithm 1 shows the overall procedure of generating negative instances. The algorithm takes input of labeled training dataset $D$ and parameter $T$ to indicate the number of instances to be generated per class.
For each class $k \in {1,2,...,K}$, it generates a set of instances, $D_k^-$, in turn, and the instances are obtained by optimizing Eq.\eqref{eq9}. 

\vspace{5pt}

\noindent{\bf Generate Seen Class Instances}\\
On the other hand, if the class $k$ is a rare class that has only a small number of samples, the trained classification model may be inaccurate. Our sample generation method can also be used to generate positive/seen class data, in order to improve the prediction accuracy. We only need to change the optimization function in step 6 of Algorithm \ref{alg:ASG} to complete this goal.

For the goal of generating a seen class instance, the loss is
\begin{equation}\label{eq11}
	\mathop{\arg\max}\nolimits_{x} P_D(x; D_k, D_k^+\cup\{x\})
\end{equation}
where $D_k^+$ is the previously generated positive data set. Again, we want to generate scattered samples, thus we force the generated instance to be distant to the previously generated ones, by the penalty:
\begin{equation}\label{eq12}
	P_3(x,D_k^+)=\max\{0,C_3-\mathop{\mathop{\arg\min}}\nolimits_{x' \in D_k^+} dist(x,x')\}
\end{equation}
where the $dist$ is a distance measure function, and $C_3$ is a positive constant.
Then the overall objective is:
\begin{equation}\label{eq13}
	\mathop{\arg\max}\nolimits_{x} P_D(x; D_k, D_k^+\cup\{x\})-\eta P_3(x,D_k^+)
\end{equation}
where $\eta$ is the coefficient of regular entry $P_3$.
Let the Eq.\eqref{eq13} replace the Eq.\eqref{eq9} in the step 6 of Algorithm 1,
the $D_k^+$ set can be obtained for each class $k$.
\vspace{5pt}

\noindent{\bf Overall Procedure}\\
After the data of unseen and seen classes are generated, it is straightforward to train a classifier for distinguishing between them. In ASG, we prefer to train such classifier for each class, i.e., train $f_k^{ooc}$ from $D_k\cup D_k^+$ and $D_k^-$.

Another issue needs to be considered is the learning capacity of the discriminator $\mathcal L$. Note that when the capacity is too high, every generated sample can be discriminated from the original data; while when the capacity is too low, the boundary of the seen classes cannot be well captured. Unlike unsupervised learning, in the OCC problem we have some data of the seen classes. Using this data, we can fine tune the hyper-parameter (e.g. the structure of a neural network) of the learning algorithm for a proper capacity.

Overall, given any learning algorithm $\mathcal L$, the procedure of the ASG framework is as follows: \\[3pt]
(1) On the seen data $D$, fine tune the hyper-parameters of $\mathcal L$; \\[3pt]
(2) Generate negative instances $D^-$ of the seen classes by Algorithm 1, as well as positive instances $D^+$ if necessary;\\[3pt]
(3) For each class $k$, train a classifier $f_k^{ooc}$ from positive data $D_k\cup D_k^+$ and negative data $D_k^-$ by $\mathcal L$;\\[3pt]
In the prediction stage,  for a test instance $x$, it is tested by each of $f_k^{ooc}$. If all $f_k^{ooc}$ classify $x$ as negative, then $x$ is predicted as a $novel$ class. Otherwise, $x$ is predicted as the class $k$ with $f_k^{ooc}$ has the highest confident.

\section{Experiments}

  \begin{figure*} [!t]
  \centering
  	\includegraphics[width=0.8\linewidth]{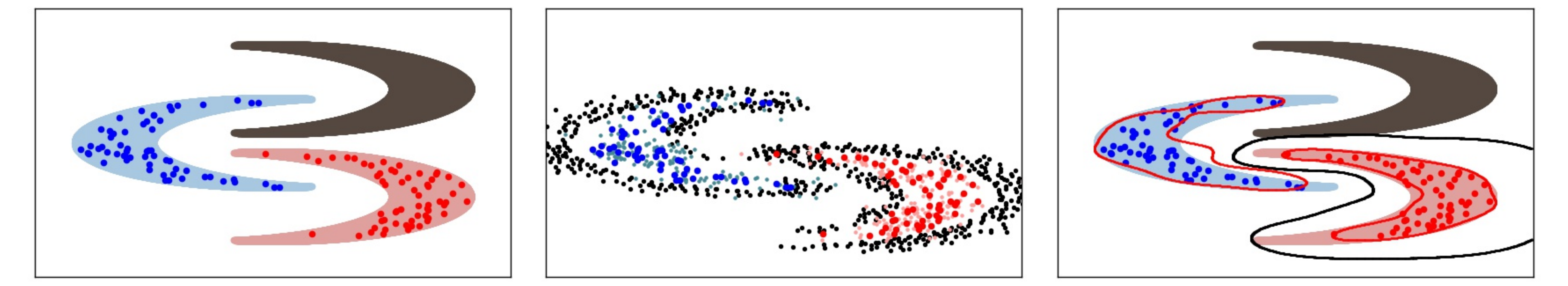}\\
  	(a) true classes and training data\ \ \ \ \ (b) original and generated data \ \ \ \ \ \ \ (c) decision boundaries\ \ \ \ \ \  \ \ \ \ 
    \caption{
    An illustration of ASG framework on the synthetic data. (a) shows the origin distribution of the three moons, where the blue and red moons are seen classes and the gray moon is unseen, and the dots are the training instances; (b) shows the generated data of seen classes (small colored dots) as well as that of unseen classes (gray dots); (c) shows the classification boundaries of a well tuned SVM (in grey line) and the ASG (in red lines). 
   }
    \label{fig1}
  \end{figure*}
  
\subsection{Comparison Methods}
As discussed above,
the hyper-parameters of the learning algorithm should be determined at the first step.
In the experiment, we employ SVM with RBF kernel as the learning algorithm and for the discriminant model too. And the binary classifier trained after the generation is also set as RBF-SVM. In order to verify the validity of the ASG framework, we conducted experiments on several benchmark datasets, comparing with:\\
\hspace*{0.02in} {\bf {OC-SVM}}:
One-Class SVM~\cite{scholkopf2001estimating} is a state-of-the-art outlier
detector~\cite{ma2003time}, which computes a binary function supposed to
capture regions in the input space where the probability density lives.\\
\hspace*{0.02in} {\bf {MOC-SVM}}:
The OC-SVM can hardly find local outliers since it seeks for a hyperplane
to separate the data and the origin in essence.
Therefore, for this comparison method, we train multiple one-class SVMs,
one OC-SVM for each seen class in the training set, for outlier detection.\\
\hspace*{0.02in} {\bf {1-vs-Set}}: 1-vs-Set Machine~\cite{scheirer2013toward}
introduces extra decision boundaries for the seen classes in order to minimize
the risk over open space.\\
\hspace*{0.02in} {\bf {OVR-SVM}}: One-vs-rest SVM is a powerful
multi-class classification scheme~\cite{rifkin2004defense}.
In the original OVR-SVM, an instance $x$ is predicted as category y, where
${y} = \mathop{\arg\max}_{k=1,...,K} f_k(x) $ where $f_k$ is a binary SVM trained for class $k$.
In order to adapt OVR-SVM for the prediction of open-category problem,
we let the $x$ be predicted as class $y$ only when $max_kf(x) > 0$,
otherwise $x$ is considered to be the instance of $novel$ class.\\
\hspace*{0.02in} {\bf {NNO}}: Nearest Non-Outlier(NNO)~\cite{Bendale15openworld} is a theoretical guaranteed method which deal with the open world recognition by introduce a measurable function in Nearest Class Mean method.

In experiments, we use the implementations of OC-SVM, MOC-SVM and OVR-SVM in the LIBSVM software~\cite{chang2011libsvm}, and the implementation of 1-vs-Set Machine from the
code released by the authors. And we implement the NNO by ourselves due to the miss of the code.
The coefficient C in SVM is determined by cross validation on the training dataset
using the original OVR-SVM.
The width of Gaussian kernel $\gamma$ is fixed to $1/d$,
where $d$ is the size of the feature.
For the ASG algorithm, the distance measure $dist$ is set to be the Euclidean distance.
When generating negative samples of class $k$,
the parameter $C_1$,$C_2$ is set as $\min_{x_1,x_2 \in D_k; x_1\neq x_2} dist(x_1,x_2)$,
$\lambda_1$ and $\lambda_2$ are both set to $0.1$, and $T=200$.
When generating positive samples for class $k$,
the parameter $C_3$ is also set to  $\min_{x_1,x_2 \in D_k; x_1\neq x_2} dist(x_1,x_2)$,
$\eta$ is set to $0.3$, and also $T=200$. For the derivative-free optimization method, we employ the recently developed RACOS algorithm\footnote{using the implementation in https://github.com/eyounx/ZOOpt}~\cite{Yu2016racos} with default parameters.

\subsection{Results}

\begin{figure*} [!b]\centering
    \rotatebox{90}{%
      \ \ Macro-averaged F1
     }
	\includegraphics[width=0.9\linewidth]{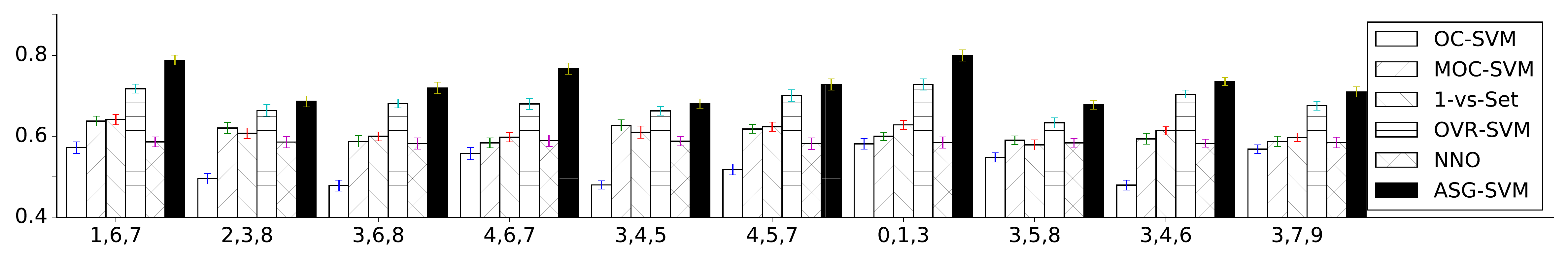}\\
	(a) with 3 seen classes \\
	\rotatebox{90}{%
      \ \ Macro-averaged F1
     }
	\includegraphics[width=0.9\linewidth]{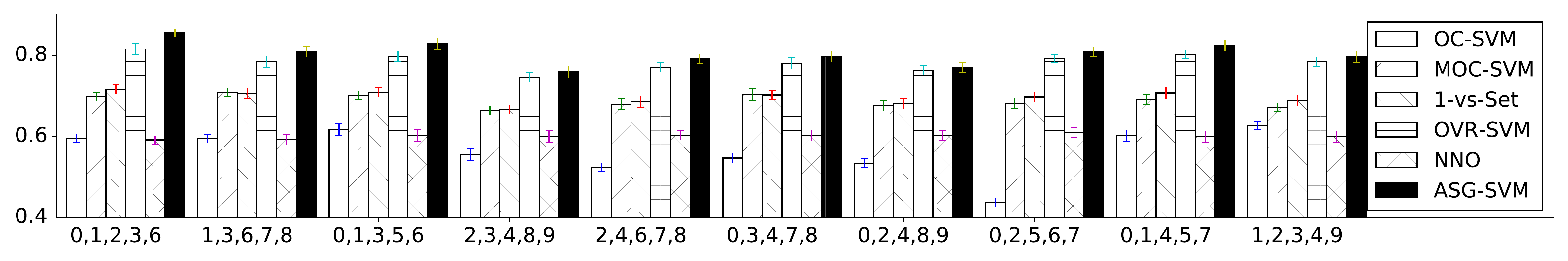}\\
	(b) with 5 seen classes
  \caption{ Comparisons of different methods on MNIST dataset}
  \label{fig2}
\end{figure*}

{\bf {Three Moons}}: We first illustrate the behaviors of ASG using a 2-dimensional synthetic data as in Figure \ref{fig1}. The dataset consists of three classes, but only two of them are seen in the training stage, as in Figure \ref{fig1} (a). With the generated data, the boundary of the seen classes can be well characterized, as in Figure \ref{fig1} (b). The decision boundaries in Figure \ref{fig1} (c) show that ASG can correctly exclude the unseen moon as seen data, while the classical SVM will classify the unseen moon as blue.

\begin{table*}[!b]
  \caption{F1, Precesion and Recall for seen classes 2,3,5 in MNIST dataset (NIPC means number of instances per class)}
  \centering
{
\begin{tabular}{c|cl|cccccc}
\hline
 & Measure	&	NIPC	&	OC-SVM	&	MOC-SVM	&	1-vs-Set	&	OVR-SVM	&	NNO	&	ASG-SVM	\\	\hline
\multirow{6}{*}{\rotatebox{90}{Seen Class}} & \multirow{2}{*}{F1}	&	100	&	.398$\pm$.033	&	.446$\pm$.028	&	.488$\pm$.046	&	.552$\pm$.021	&	.424$\pm$.013	&	\textbf{.570$\pm$.029}	\\	
&	&	1000	&	.389$\pm$.026	&	.437$\pm$.027	&	.542$\pm$.043	&	.612$\pm$.013	&	.421$\pm$.010	&	\textbf{.624$\pm$.024}	\\	\cline{2-3}
& \multirow{2}{*}{Precision}	&	100	&	.330$\pm$.037	&	.414$\pm$.033	&	.336$\pm$.051	&	.398$\pm$.019	&	.367$\pm$.012	&	\textbf{.539$\pm$.027}	\\	
&	&	1000	&	.314$\pm$.027	&	.374$\pm$.031	&	.388$\pm$.048	&	.534$\pm$.017	&	.379$\pm$.011	&	\textbf{.566$\pm$.022}	\\	\cline{2-3}
& \multirow{2}{*}{Recall}	&	100	&	.502$\pm$.031	&	.484$\pm$.030	&	.882$\pm$.046	&	\textbf{.898$\pm$.024}	&	.501$\pm$.014	&	.605$\pm$.031	\\	
&    &	1000	&	.511$\pm$.029	&	.525$\pm$.027	&	\textbf{.897$\pm$.044}	&	.716$\pm$.014	&	.474$\pm$.012	&	.697$\pm$.025	\\	\hline
\multirow{6}{*}{\rotatebox{90}{Unseen Class}} & \multirow{2}{*}{F1}	&	100	&	.645$\pm$.035	&	.918$\pm$.031	&	.882$\pm$.048	&	.895$\pm$.023	&	.702$\pm$.012	&	\textbf{.933$\pm$.030}	\\	
&	&	1000	&	.617$\pm$.031	&	.905$\pm$.029	&	.897$\pm$.044	&	.930$\pm$.016	&	.722$\pm$.011	&	\textbf{.935$\pm$.026}	\\	\cline{2-3}
& \multirow{2}{*}{Precision}	&	100	&	.736$\pm$.036	&	.941$\pm$.034	&	.983$\pm$.049	&	\textbf{.987$\pm$.020}	&	.763$\pm$.014	&	.957$\pm$.028	\\	
&	&	1000	&	.726$\pm$.027	&	.944$\pm$.033	&	\textbf{.986$\pm$.043}	&	.968$\pm$.018	&	.763$\pm$.012	&	.966$\pm$.024	\\	\cline{2-3}
&  \multirow{2}{*}{Recall}	&	100	&	.577$\pm$.032	&	.897$\pm$.032	&	.752$\pm$.053	&	.818$\pm$.025	&	.650$\pm$.013	&	\textbf{.911$\pm$.034}	\\	
& &	1000	&	.538$\pm$.030	&	.870$\pm$.028	&	.789$\pm$.048	&	.896$\pm$.013	&	.685$\pm$0.01	&	\textbf{.907$\pm$.026}	\\	\hline
\end{tabular}}
\label{table_1}
\end{table*}

\begin{table}[t!]
\caption{Improvement ratio of ASG-SVM to the comparing methods on MNIST dataset (NIPC means number of instances per class)}
\label{table_4}\centering
\resizebox{\linewidth}{!}{
  \begin{tabular}{l|c@{\ \ }c@{\ \ }c@{\ \ }c@{\ }c}
  \hline
  \#classes & \multicolumn{4}{c}{Comparison Methods} \\ \cline{2-6}
  -NIPC  & OC-SVM  & MOC-SVM  & 1-vs-set & OVR-SVM  & NNO\\ \hline
  3-1000 & .2391  & .1720   & .1782   & .0624  & .1624 \\
  3-100  & .3816  & .2061   & .1957   & .0650  & .2469\\
  5-1000 & .3995  & .0755    & .1016   & .0108  &  .3877\\ 
  5-100  & .4283  & .1691   & .1552   & .0259  & .3403\\\hline
\end{tabular}}
\end{table}

\hspace*{0.02in} {\bf {Handwritten Digit Image Classification}}:
We conducted the second experiment on the MNIST handwritten digit dataset. First, seen categories were randomly selected from 10 total categories and the test data contains all 10 categories. The number of training data, generated data and test data are 100, 300 and 100 for each class. We have 10 random selections for the seen categories and repeat 10 times for each selection, both mean and standard deviation are recorded. Figure~\ref{fig2} shows the results on 3 and 5 seen classes in training. It can be observed that ASG-SVM achieves the best performance on both configurations, 
while the OVR-SVM is the runner up. Besides, the 1-vs-set machine shows to outperform the OC-SVM and MOC-SVM, 
and the outlier detection methods, OC-SVM and MOC-SVM, produce the worst performance. Besides, the NNO seems not perform well in this dataset. 

To break down the performance to seen and unseen data separately, we set the class $2,3,5$ as seen categories, and training data size for each category is set to 100 and 1000, respectively. The experiments are repeated for 10 times, and both mean and standard variance of the measure are reported. The results of F1, precision and recall for both seen classes and unseen classes are shown in Table~\ref{table_1}. For seen classes, the ASG-SVM gets the best performance on F1 and precision when the instance number for each category is set to 100 and 1000,  while the recall of OVR-SVM and 1-vs-set is the highest when the number of  instance is set to 100 and 1000. For unseen classes, the ASG-SVM obtains the highest performance on F1 and recall, while the precision of OVR-SVM and 1-vs-set is the best when the instance number of each class is 100 and 1000.

Table~\ref{table_4} shows the improvement ratio of ASG-SVM over the comparing methods in 4 configurations of the number of seen classes and training set size. The cases in the first column represent different situations, for example, 3-100 represent there are 3 seen classes in the training data and 100 instances for each class. It can be observed that, when the training data has a smaller size, ASG-SVM has larger improvements. This mainly due to that ASG also generates samples of seen classes to improve the prediction accuracy.
\vspace{6pt}

\noindent {\bf {Document Classification}}: 
In the third experiment, we conduct experiments on the 20 Newsgroups dataset, which is a popular text dataset consists of documents from 20 different topics. 
Note that this data set has some topics that are highly similar, such as $comp.sys.ibm.pc.hardware$ and $comp.sys.mac.hardware$. When one belongs to the seen classes and the other is unseen, the classification will be quite difficult. We set the number of seen classes as 5. The number of training data, generated positive data, generated negative
data and test data are set to 3000, 2000, 2000 and 3000, respectively.
We randomly sample seen classes and the training data to form 20 data sets, and for each data sets,
the experiment is repeat for 10 times.
%The results of $win/tie/loss$ counts of all possible pairs of methods are shown in Table~\ref{table_2}. The number in the $i$-th row and the $j$-th column indicates the times of win, tie and loss of the method in the corresponding row, compared to that in the corresponding column, over all the 20 configurations.

\begin{figure} [!h]\centering
    \rotatebox{90}{%
      \ \ \ \ \ \ \ \ \ \ \  Macro-averaged F1
     }
	\includegraphics[width=0.8\linewidth]{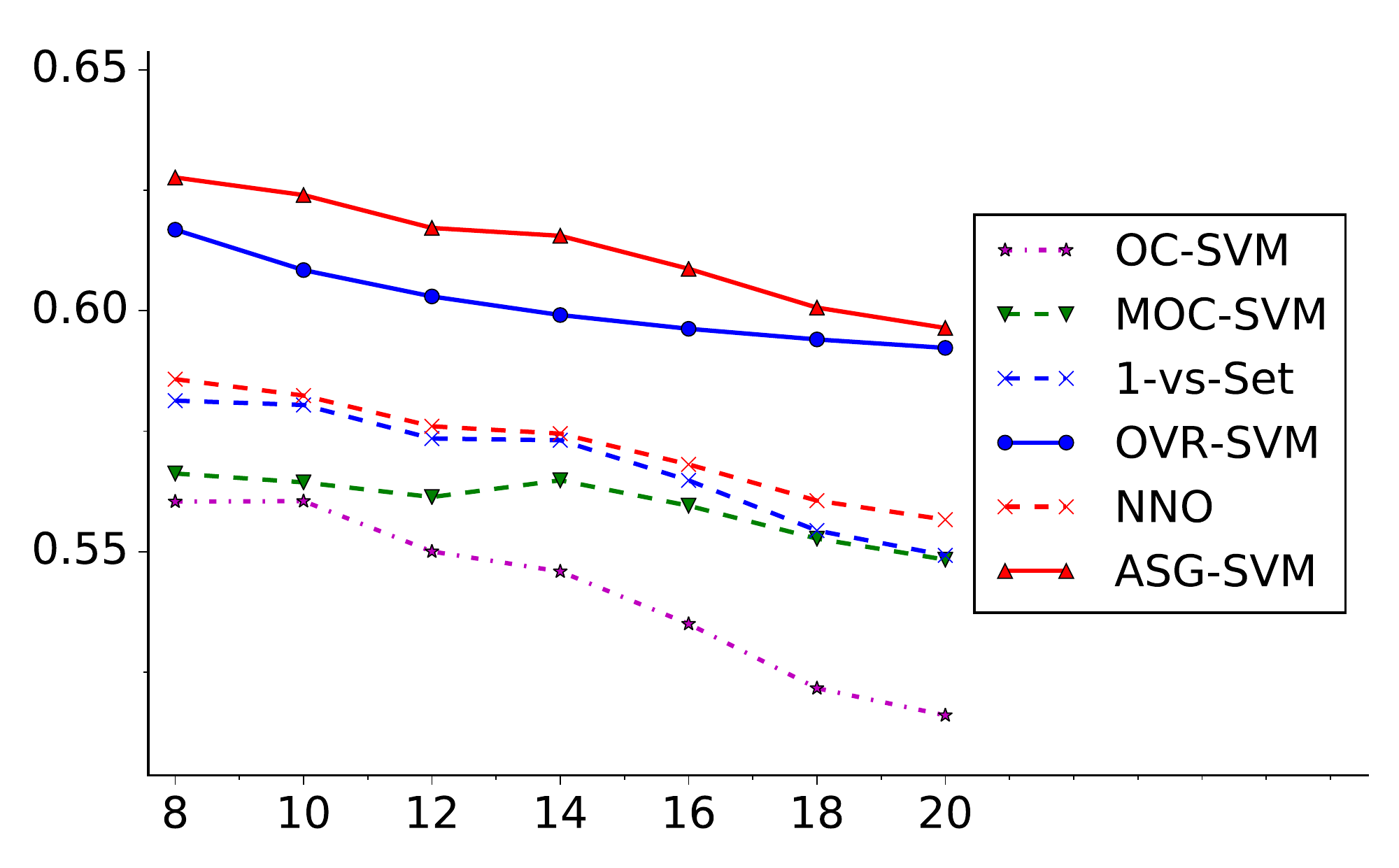}\\
	Number of test classes \ \ \    
  \caption{ Performance with different number of test classes on 20News dataset}
  \label{fig5}
\end{figure}

%The data in the Table~\ref{table_2} shows that, the proposed ASG framework wins all the configurations compared with the OC-SVM,  MOC-SVM and outperforms the OVR-SVM and 1-vs-Set machine with the loss rate 5\%. Except for ASG-SVM, OVR-SVM wins all the comparisons, including 1-vs-Set machine and NNO, and the latter two methods has a similar performance in this dataset. Besides, the outlier detection methods lose all the comparisons. 
Figure~\ref{fig5} shows the Macro-F1 performance with the increase of test classes. It can be observed that the curves in general decreases as expected. Among these methods, ASG-SVM consistently achieves the best performance, and the OVR-SVM is the runner-up. The outlier detection methods, OC-SVM and MOC-SVM, fail in this data, which further confirms that the instance of unseen classes should not be simply treated as outliers in the OCC problem. The win/tie/loss table on all algorithm pairs is in the appendix due to the space limit, which will show that  ASG framework is significantly better than the other methods.

\begin{table}[!t]
\centering
\caption{Comparisons of classification performance (Macor-F1 score, mean$\pm$std.), the best performance on each dataset is in bold.}
\label{table_3}
\resizebox{\linewidth}{!}{\small
\begin{tabular}{@{}c@{\ }|@{\ }c@{\ }c@{\ }c@{\ }c@{\ }c@{\ }c@{}}
\hline
Dataset &	artificial	& flag	& glass	& letter	& page\_blocks \\\hline
OC-SVM	&.419$\pm$.036	& .309$\pm$.052	& .482$\pm$.039	& .658$\pm$.024	& .505$\pm$.049 \\			
MOC-SVM	&.555$\pm$.022	& .504$\pm$.050	& .678$\pm$.037	& .721$\pm$.019	& .536$\pm$.042 \\		
1-vs-Set	&.562$\pm$.042	& .462$\pm$.057	& .609$\pm$.046	& .739$\pm$.030	& .513$\pm$.046	\\
OVR-SVM	&.667$\pm$.020	& .501$\pm$.042	& .646$\pm$.032	& .885$\pm$.015	& .579$\pm$.037 \\
NNO	&.564$\pm$.013	&\textbf{.581$\pm$.045}	&.659$\pm$.014	&.938$\pm$.016	&.519$\pm$.036 \\	
ASG-SVM	& \textbf{.703$\pm$.035}	& .522$\pm$.043	& \textbf{.706$\pm$.042}	& \textbf{.944$\pm$.026}	& \textbf{.611$\pm$.043}
\\ \hline
\end{tabular}}
\end{table}
% \usepackage{multirow}
\vspace{6pt}

\noindent {\bf {Small Datasets}}:
We also conducted experiments on five small datasets,
where the size for each seen class is no more than 100 in the training datasets,
which makes the classification tasks become more challenging.

As the result shown in  Table~\ref{table_3}, ASG-SVM obtains the best performance in four datasets compared to the other five methods; NNO has the best performance in the $flag$ dataset, but can be worse than some of the comparing methods. In addition to ASG-SVM,  NNO and OVR-SVM are runner-ups that have better performance in several cases, but have lower Macro-F1 in the $glass$ dataset than MOC-SVM, 
where the number of each class is just 15 in the training set. Sort the algorithms with their average ranks on all data sets, the best is ASG-SVM (average rank 1.2), and the remaining are NNO (2.6), OVR-SVM (3.0), MOC-SVM (3.6), 1-vs-Set (4.6), and OC-SVM (6).

\section{Conclusion}
  Open-category classification problem often occurs in practical problems, where a system needs to predict the data in an open environment. In this paper, we propose the ASG framework to address the problem by adversarial data generation. In experiments, we demonstrate that ASG can successfully generate boundary data around the seen classes, which makes the recognition of unseen classes can be easily done by supervised learning. On several datasets, ASG shows to be more effective than several state-of-the-art methods. In the future, besides the current SVM model, we would like to apply ASG with state-of-the-art multi-class learning method, and develop a theoretical grounded method for the OCC problem.

\bibliographystyle{named}
\bibliography{asg}

\section*{Appendix}

\begin{table}[!h]
\caption{The results of win/tie/loss information with 20 randomly selected
configurations on 20Newsgroup dataset
(paired two-tailed t-test at 95\% significance level)}
\label{table_2}\centering
{
\begin{tabular}{l|llllll}
\hline
Method   & OC-SVM & MOC-SVM & 1-vs-Set & OVR-SVM & NNO   & ASG-SVM \\ \hline
OC-SVM   & -      & - & -   & -  &-  & -  \\
MOC-SVM  & 15/5/0 & -       & -   & -  & - & -  \\
1-vs-Set & 20/0/0 & 17/2/1  & -        & -  & -   & -  \\
OVR-SVM  & 20/0/0 & 20/0/0  & 20/0/0   & -  & -  & -  \\
NNO      & 16/4/0 & 12/3/5  & 9/8/3    & 0/0/20  & - & -  \\
ASG-SVM  & 20/0/0 & 20/0/0  & 19/0/1   & 16/3/1  &20/0/0  & -  \\ \hline
\end{tabular}}
\end{table}

\end{document}